\documentclass[twocolumn]{article}

\usepackage{framed,multirow}
\usepackage{fancyhdr}
\usepackage[includefoot,includehead,top=2.5cm,bottom=2.5cm,left=2.5cm,right=2.5cm]{geometry}
\usepackage{graphicx}
\usepackage{natbib}

\usepackage{amssymb}
\usepackage{latexsym}
\usepackage{url}
\usepackage{xcolor}
\definecolor{newcolor}{rgb}{.8,.349,.1}

\usepackage{diagbox}
\usepackage{tabularx}

\fancypagestyle{plain}{%
	\fancyhf{}%
	\fancyfoot[L]{\copyright  2020. This manuscript version is made available under the CC-BY-NC-ND 4.0 license http://creativecommons.org/licenses/by-nc-nd/4.0/}%
	\fancyfoot[R]{\thepage}%
}

\title{Multi-task learning for natural language processing in the 2020s: where are we going?}

\author{Joseph Worsham\\ University of Colorado Colorado Springs
\and Jugal Kalita\\ University of Colorado Colorado Springs}

\date{}

\begin{document}
	
\maketitle

\begin{abstract}
Multi-task learning (MTL) significantly pre-dates the deep learning era, and it has seen a resurgence in the past few years as researchers have been applying MTL to deep learning solutions for natural language tasks. While steady MTL research has always been present, there is a growing interest driven by the impressive successes published in the related fields of transfer learning and pre-training, such as BERT, and the release of new challenge problems, such as GLUE and the NLP Decathlon (decaNLP). These efforts place more focus on how weights are shared across networks, evaluate the re-usability of network components and identify use cases where MTL can significantly outperform single-task solutions. This paper strives to provide a comprehensive survey of the numerous recent MTL contributions to the field of natural language processing and provide a forum to focus efforts on the hardest unsolved problems in the next decade. While novel models that improve performance on NLP benchmarks are continually produced, lasting MTL challenges remain unsolved which could hold the key to better language understanding, knowledge discovery and natural language interfaces.
\end{abstract}

\section{Introduction}
Multi-task learning (MTL) is a collection of techniques intended to improve generalization, strengthen latent representations and enable domain adaptation within the field of machine learning \citep{mtl97}. It has been applied to feed-forward neural networks \citep{mtl97}, decision trees \citep{mtl97}, random forests \citep{mtforest08}, Gaussian Processes \citep{mtgp08}, support-vector machines \citep{mtsvm04} and, most recently, deep neural networks \citep{mtloverview17} across a broad range of domains. This includes specific deep learning architectures such as MTL seq2seq models \citep{mtlseqseq18} and MTL transformers \citep{mtdnn19}. It has been shown that under certain circumstances, and with well-crafted tasks, MTL can help models achieve state-of-the-art performance on a range of different tasks \citep{whichtask19}. It has also been shown, however, that MTL can be extremely fragile and sensitive to both the selected tasks and the training process which leads to models that significantly under-perform when compared to the best single-task models \citep{whenmtl17}. While MTL has been a subject of research for multiple decades \citep{mtloverview17}, there still exist a number of unsolved problems, unexplored questions and shortcomings in production systems which are addressed within. This survey will present a condensed summary of the large library of current MTL research applied to natural language processing (NLP) and present a set of goals intended to help highlight the MTL problems that we should strive to solve in the next decade.

\section{Characterizing Multi-Task Learning}
MTL introduces additional training objectives to a learning system to bias the learner with a broader understanding through solving related tasks. The end-goal is to improve performance on a set of primary tasks through the inductive bias introduced by the additional tasks \citep{mtl97}. The set of primary tasks are referred to as the target task set, and additional tasks, which are used to improve performance on the target set, belong to the auxiliary task set. While this is the standard approach \citep{mtloverview17}, others have also designed MTL models with no auxiliary focusing on competitively solving all the tasks jointly \citep{decanlp18}.

\begin{figure*}
	\begin{center}
		\includegraphics[width=\textwidth,height=6.2cm,keepaspectratio]{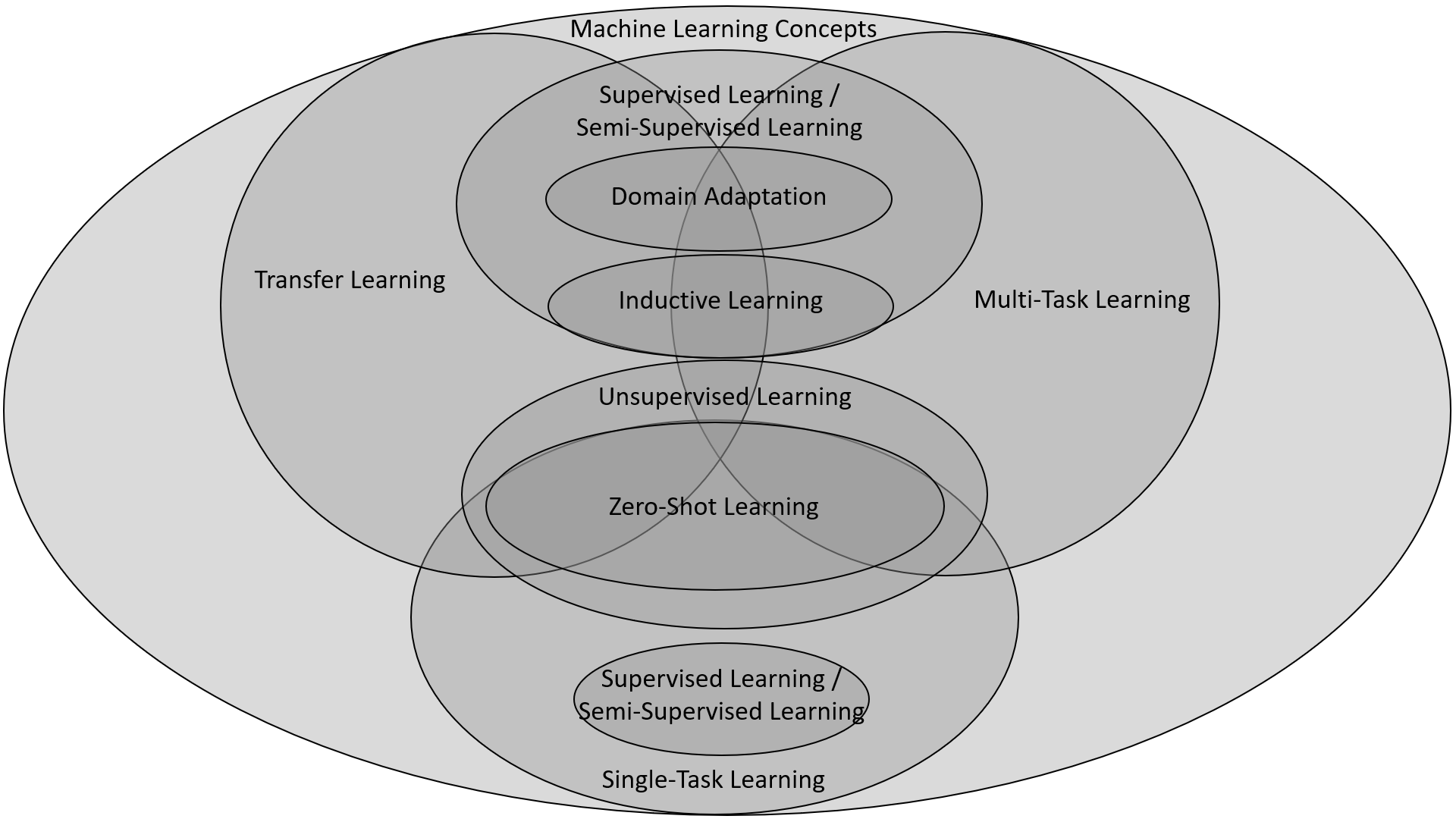}
		\caption{Relationship of Machine Learning Concepts with a Focus on Transfer Learning and Multi-Task Learning}
		\label{ml_concepts}
	\end{center}
\end{figure*}

In practice, MTL is closely related to Transfer Learning (TL) \citep{learn2learn98}, as the goal of each is to improve the performance of a target task or domain through the use of related tasks and domains. A task is defined as a specific operational capability, such as Part of Speech Tagging or Question Answering. Tasks traditionally do not share the same output features or solution space. A domain is a certain feature space and underlying data generating process. When working with TL and MTL, commonly, different domains share the same feature space, but have different data generating processes. While there is no limit on how much variance can exist in different but related domains, a common example in NLP is to treat different languages as different domains \citep{tlmt16}. Both TL and MTL can make use of differing domains and differing tasks.
	
Transfer Learning is broken down into three different categories based on what differs between the source and the target \citep{advancedda19}. If the source and target share the same task with different domains, this is called Transductive Transfer Learning, commonly known as Domain Adaptation. If the source and target share the same domain with different tasks this is Inductive Transfer Learning which learns to transfer via inductive bias. If the source and target have different domains and different tasks this is a form of unsupervised Transfer Learning which learns common representations despite having no direct similarities between the source and the target.

With this TL taxonomy, we formulate a related breakdown for MTL. While MTL terminology is traditionally focused on varying tasks it is also possible to train jointly on different domains. If the source task and auxiliary tasks are the same with different domains, we label this Transductive Multi-Task Learning, or Domain Regularization. When the source task and auxiliary tasks are different but share the same domain this is the standard form of MTL which we formally identify as Inductive Bias MTL. Finally, if the source and auxiliary tasks are different and do not share the same domain we call it Multi-Task Feature Learning, originally introduced by \cite{unrelatedtasks12} Table \ref{mtl_taxonomy} shows this breakdown for both TL and MTL.

A relational representation of these concepts is provided in Figure \ref{ml_concepts}. This shows that TL and MTL share some overlap, indicating that these techniques can be used together. While \cite{whichtask19} show that there are significant differences in the types of tasks proven to be useful in MTL vs. TL, \cite{bentask17} argue that they produce similar observed benefits. \cite{mtdnn19} show that TL and MTL are complementary to one another when used in combination to train a complex model, which is in contradiction to earlier work that showed, in different circumstances, this combination yielded no significant improvement \citep{howtrans16}. These techniques also overlap with standard single-task learning through a method called zero-shot learning. Zero-shot learners, or generalist agents, are capable of jointly understanding many tasks or concepts with no fine-tuning on specific tasks \citep{decanlp18}.

\begin{table}[t]
	\caption{TL / MTL Task and Domain Categories}\smallskip
	\centering
	\resizebox{.95\columnwidth}{!}{
		\smallskip\begin{tabular}{|l|l|l|}
			\hline
			& Same Tasks & Different Tasks\\ \hline
			Same Domains & Standard Learning Setting & \diagbox[dir=SW]{Inductive\\Transfer\\Learning}{Inductive Bias\\(standard) MTL}\\ \hline
			Different Domains & \diagbox[dir=SW]{Transductive\\Transfer Learning\\(Domain Adapation)}{Transductive MTL\\(Domain Reuglarization)} & \diagbox[dir=SW]{Unsupervised\\Transfer Learning}{Multi-Task\\Feature Learning}\\ \hline
		\end{tabular}
	}
	\label{mtl_taxonomy}
\end{table}

\section{When to Use Multi-Task Learning}
One of the biggest needs for a successful machine learning system is access to extremely large amounts of labeled data. MTL is proposed as a technique to help overcome data sparsity by learning to jointly solve related or similar problems to produce a more generalized internal representation. Regardless of the number of target tasks to solve, MTL can only be considered useful when at least one target task is improved upon when compared to a collection of single-task models \citep{whichtask19}.

Along with enabling zero-shot learning \citep{decanlp18}, MTL is commonly presented as a regularization technique to aid in the generalization of a task to unseen examples \citep{mtl97,mts2sl15,mtdnn19,gpt19}. This belongs to the Transductive MTL class in Table \ref{mtl_taxonomy}.

Additionally, MTL has desirable traits when it comes to efficiency. \cite{bertandpals19} describe a well designed MTL model to be computationally less complex, to have fewer parameters limiting the burden on memory and storage and to ultimately require less power consumption at inference time. \cite{whichtask19} also point out the desirable trait of quicker inferencing depending on the architecture of the MTL model. These characteristics pertain to neural network MTL implementations which are less expensive than deploying a complete model for each class individually.

\section{MTL Implications and Discoveries}
Researchers have been studying the implications and nuances of MTL when compared to traditional single-task training since its introduction. Given the human intuition of how MTL can help improve model performance, practitioners are often surprised at how delicate and sensitive these algorithms can be \citep{whenmtl17,whichtask19}. This section will discuss MTL discoveries in this regard through the topics of task relationship, dataset diversity, model design considerations and training curriculum. Techniques identified in this section are shown relationally in Figure \ref{ml_techniques}.

\subsection{Task Selection}
The similarities between a set of tasks are commonly cited as one of the most influential design factors in building MTL systems. Through a series of experiments, \cite{mtl97} showed that the benefit of MTL is due to the direct knowledge learned from the auxiliary tasks. He further showed that some inductive bias can actually harm performance. A host of other researchers have gone on to argue that task relatedness plays a key role in determining how knowledge information is shared \citep{howtrans16,notion08} and when an auxiliary task will help and when it will hurt \citep{whichtask19,probing19}.

\cite{whichtask19} continued to explore this concept and showed that tasks which seem related can often have underlying complex and competing dynamics. They present a table showing how every factorized task pair performed in relation to a single-task model, and, via the hypothesis that related tasks improve performance, show which tasks they believe to be related. While this work was not performed on a set of NLP tasks, it showed the importance of task relationship and provided a novel way to measure relatedness. 

Another unique observation published by \cite{whichtask19} shows that tasks which are beneficial in a TL environment appear to perform poorly as auxiliary tasks in an MTL setting. This raises a question on whether or not dissimilar tasks could be used intentionally to help regularize a system or even protect a system from adversarial attack. While this observation seems to disagree with the conventional wisdom that only similar or related tasks can lead to improved model performance \citep{notion08}, they are not alone. \cite{unrelatedtasks12} have also shown that unrelated tasks can still be beneficial. This poses a unique opportunity to further explore task relationships and usefulness on the most recent MTL benchmarks.

\subsection{MTL Dataset Considerations}
The datasets used for target tasks and auxiliary tasks play an important role in building successful MTL systems. The first topic addressed considers how the size and diversity of the datasets impact the learning of a model. \cite{mts2sl15} perform a set of experiments to determine how the size of the datasets for the target and auxiliary tasks impact the overall results of the model on the target set. They show that the size ratio between a target task dataset and an auxiliary task dataset does have an impact on performance. They argue that when the target dataset is large, the best MTL performance is achieved with a small auxiliary dataset with a size ratio between 0.01 and 0.1. When this mixing ratio gets too high it is shown that the model will overfit to the auxiliary task at the cost of performance on the target task. Other researchers agree that the best performance is achieved with a small number of auxiliary task updates compared to target task updates \citep{whenmtl17,probing19} and that adding more data to a poorly selected auxiliary task can significantly harm the model \citep{probing19}.

Researchers have also considered the underlying properties and statistics to determine how they impact MTL performance. A theoretical definition of MTL and task relatedness is presented by \cite{notion08}. The goal of this work is to develop a formulated approach for determining when MTL is advantageous and to what degree. They seek a theoretical justification for task relatedness based on measurable similarities found within the underlying data generating processes of each task. While their definition of task more closely relates to the definition of a domain within this survey, they establish formal error bounds to measure and learn task relatedness.

Recent work has gone on to argue that size is not a useful metric for determining MTL gain \citep{bentask17}. Research has shown that simple tasks, requiring few training iterations, and difficult tasks, which struggle to converge on a solution, do not lead to the development of useful MTL representations \citep{mtl97,decanlp18}. \cite{whenmtl17} argue that MTL task selection should be addressed via data properties, not intuition on what a human performer may consider easy. They perform a set of studies that measure statistical distributions of supervised labels in auxiliary task datasets and find that the best performance is achieved when the auxiliary tasks have compact mid-entropy distributions. That is to say, the best auxiliary tasks are neither too easy to predict nor too difficult to learn.

Another perspective on underlying properties of the auxiliary datasets is to consider the loss produced by each task while learning. The magnitude of the task loss can be considered a task similarity metric. \cite{whichtask19} show that imbalanced tasks in this regard produce largely varied gradients which can confuse model training. Oftentimes, task selection is not something that can be changed, but \cite{whichtask19} recommend using a task weighting coefficient to help normalize the gradient magnitudes across the tasks. Similar to task loss, the learning curve, showing how loss decreases over training, is also proposed as a metric for task similarity. It was found that MTL gains are more likely when a target task's learning curve plateaus early in training and the auxiliary tasks do not plateau \citep{bentask17}.

It is also worth noting that Data Augmentation is a proven technique to help overcome data sparsity and improve task performance in TL and MTL settings. \cite{dataaugment19} propose LAMBADA, a language generator model fine-tuned on a small task-specific dataset, which generates semi-supervised training data to augment the data available to a task specific language classification task.

\subsection{Model Selection and Design}
\label{model_design}

\begin{figure*}
	\begin{center}
		\includegraphics[width=\textwidth,height=6.28cm,keepaspectratio]{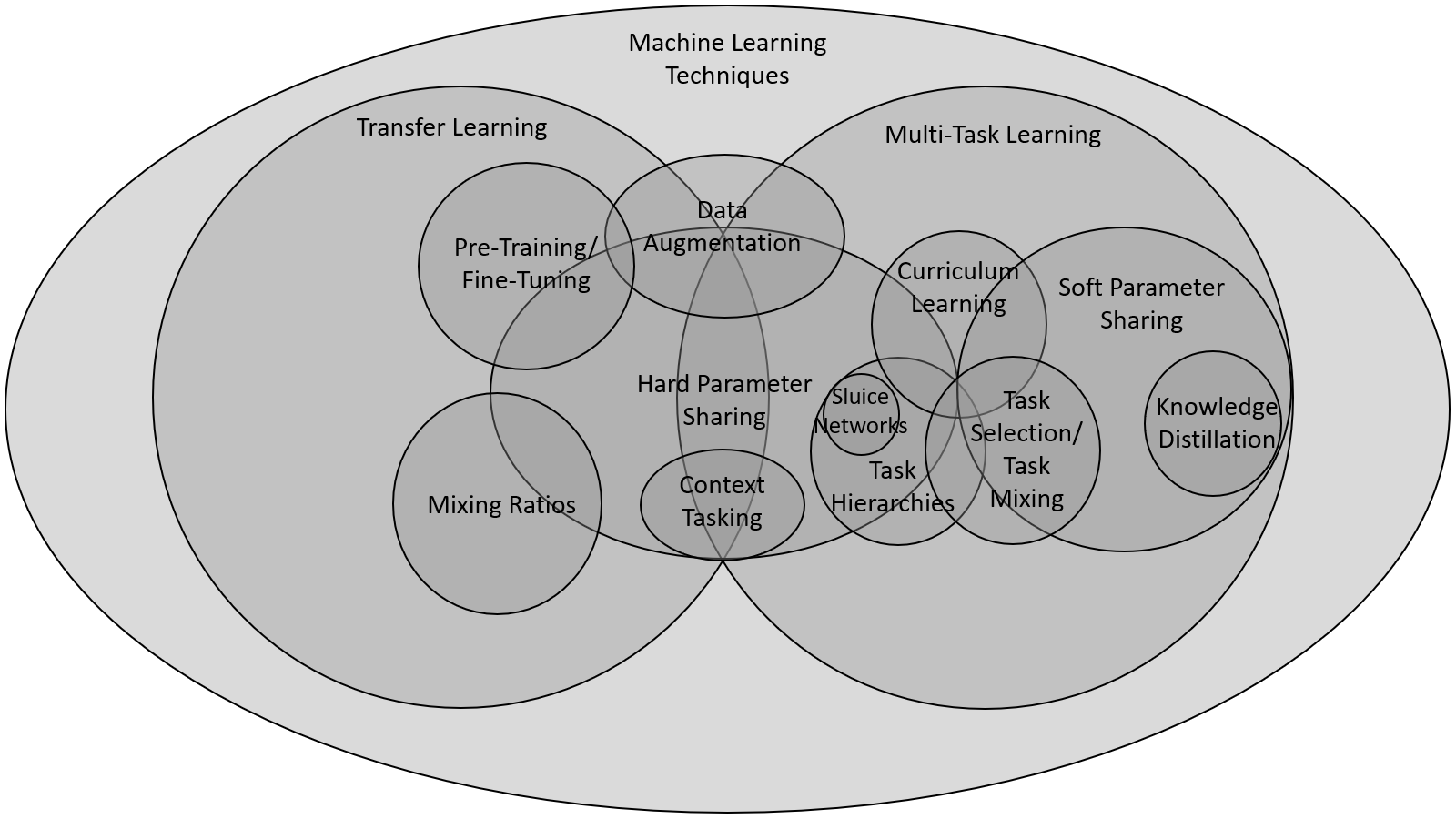}
		\caption{Relationships of Transfer Learning and Multi-Task Learning Techniques for Deep Learning}
		\label{ml_techniques}
	\end{center}
\end{figure*}

There is a large body of research considering how MTL influences model selection and design. While the authors acknowledge the importance of other machine learning models, this section will focus solely on neural networks due to recent deep learning trends. Figure \ref{ml_techniques} shows the primary MTL techniques for Deep Learning and their relationships. There are two primary families of MTL neural network approaches: hard parameter sharing and soft parameter sharing \citep{mtloverview17}. Hard parameter sharing is the approach most closely related to traditional machine learning techniques and the same mechanism used by many transfer learning solutions. In hard parameter sharing, full network layers and their parameters are shared directly between tasks. Soft parameter sharing is more specialized and instead creates separate layer parameters for each task. These task-specific layers are then regularized during training to reduce the differences between shared layers. This encourages layers to have similar weights but allows each task to specialize specific components. Diagrams showing these network concepts can be found in the MTL survey by \cite{mtloverview17}. Knowledge Distillation (KD) is a proven soft parameter technique which trains a student network to imitate the full output of a trained teacher network \citep{kd15}.

An early study in MTL showed that auxiliary tasks which help increase performance on a target task prefer to share hidden layers and weights with the target task, while unhelpful auxiliary tasks prefer to use weights not used by the target task \citep{mtl97}. This intuition has laid the groundwork for deep learning models which focus on building an enhanced internal representation of a problem space through shared hidden layers. It has been shown that pre-defining which layers to share can improve the performance of a deep learning MTL model when the tasks are generally beneficial, but this can break down if the wrong task pairs are selected \citep{mtloverview17}. The study goes on to argue for the benefit of learning task hierarchies internal to the model during training to help overcome this problem. Research has also shown that the depth of a layer and the benefit of sharing the layer between two tasks can be considered a measure of similarity of the two tasks \citep{howtrans16}. They argue that low-level layers, such as word embeddings, are generally useful for all NLP tasks, while higher level layers become more specific and can only be shared among more similar tasks. This suggests that model architectures can be built off this metric when combined with other evaluations of task relatedness.

Another model consideration when building MTL systems is the capacity of the network. \cite{gpt19} prove that the capacity of a language model is essential to good performance and that increasing capacity produces a log-linear improvement. This follows conventional neural network wisdom and agrees with other research, such as BERT \citep{bert19}, whose performance appears to scale with the model size, and T5 \citep{t519}, which achieved state-of-the-art results when pushed to 11 billion parameters.

\cite{sluice17} show that hard-parameter sharing, task-specific network layers, hierarchical NLP layers and a regularizer to encourage tasks to share only what is useful, called block-sparse regularization, can be combined to create a powerful MTL network called a Sluice network. The Sluice network consistently performed better than single-task multi-layer perceptrons (MLPs) on all evaluation tasks and outperformed traditional hard parameter sharing approaches \citep{mtl97} on most NLP tasks.

An additional question that must be addressed is how a task is represented within the model. It is common with Inductive Bias MTL that each task has a specific set of output layers that can be queried to return task specific results \citep{mtloverview17}. However, \cite{decanlp18} present a novel idea in which the task itself in included as input to the network, identified within this survey as Context Tasking. While the implementation may differ across domains and tasks, Context Tasking was implemented here by representing each task as a natural language question with a natural language answer. This avoids the need for any task-specialized components and naturally supports zero-shot learning and open-set classification \citep{openset14}. \cite{ellipsis19} present another interesting approach to Context Tasking by casting the NLP tasks of ellipsis resolution and coreference resolution as reading comprehension problems and produced new state-of-the-art results using Inductive Bias MTL.

\subsection{Training Curriculum}
A final topic of MTL training implications is the design of a training curriculum. Given the research above regarding mixing ratios, task weighting, shared representations and sensitivities to task selection, it seems natural that MTL should be addressed with an intelligent training curriculum. The standard curriculum for MTL tasks is to build mini-batches containing examples for a single task and then alternate between tasks during training. The ratio of task mini-batches can be identical for all tasks or varied based on task performance or dataset size \citep{mtl97,mts2sl15,mtloverview17,glue18}. \cite{decanlp18} refer to this as a fixed-order round robin curriculum and prove that it works well on tasks that require few iterations, but it struggles with more complex tasks. They furthermore consider hand-crafted curricula and show that beginning with more difficult tasks and slowly introducing additional tasks performs the best. Other work has considered including task-specific stopping conditions for TL and MTL \citep{howtrans16}, and more recent research has proposed a teacher-based annealing solution to dynamically control the auxiliary task impact with KD \citep{bam19}. Other research has shown that round robin training is most impactful towards the end of the curriculum \citep{bertandpals19}. They propose a technique called annealed sampling in which batch sampling is originally based on the ratio of dataset sizes and slowly anneals to an even distribution across all tasks as the current epoch number increases. These discoveries, when combined with curriculum research emerging from the field of reinforcement learning \citep{curriculum17}, lead to a wealth of new research opportunities towards the design of MTL curricula.

\section{Learning Task Relationships}
Beyond the research into measuring properties of tasks and datasets to determine similarities, there have been multiple efforts to intrinsically learn task relatedness through a learning process. \cite{mtl97} showed that neural networks trained in an MTL setting exhibited a behavior where related tasks would share hidden nodes and unrelated tasks would not. This discovery implies that neural networks are able to determine what information is useful for sharing between tasks without an explicit signal conveying the task relationship. It is therefore reasonable to believe that neural networks are able to learn, and even describe, task relationship explicitly through the MTL training process. Research has since explored different clustering techniques built on this discovery which attempt to cluster network weights and parameters leading to a latent task relationship embedded in the task clusters \citep{mtloverview17}. Not only do these techniques inherently learn task relationships, they also help to train neural networks by penalizing them from diverging too much from a common set of knowledge shared by similar tasks. \cite{mtloverview17} also presents the Deep Relationship Network and the Cross-Stitch Network which are hard and soft parameter models, respectively, able to identify task relationship through training.

An approach called task2vec has been proposed which learns an embedding vector for an entire task that is agnostic to the size of the dataset \citep{task2vec19}. The embedding attempts to capture semantic similarities between tasks by training a model to solve a task, and then probing the network to approximate the amount of information carried by the weights. The proximities between two task embedding vectors are theorized to represent task relatedness while the magnitude of the embedding vector is thought to correlate to the complexity of the task.

An alternative approach to learning directly from the hidden nodes and gradients is to efficiently search through task pairs to determine task similarities. Depending on the number of tasks an exhaustive search very quickly becomes impossible, however, heuristic based searches have been found to act as a good stand-in to estimate when tasks may be related \citep{whichtask19}. They show that there is a high correlation between the validation loss of a network trained on 20\% of the data and the fully trained validation loss of a network. Based on this claim, they use the loss at 20\% as a heuristic to lightly train multi-task permutations for finding optimally performing task sets. They go on to show that given three tasks, the average loss of every two-pair combination is an effective approximation of the loss when all three tasks are trained jointly. This acts as a good search heuristic for finding optimized task sets. While this work has focused on small task sets and relatively small combinations, others have shown the benefit of having many auxiliary tasks to boost MTL performance \citep{mtloverview17,snorkel18,mtdnn19}. More research into the implications of these findings is important to understanding the effect of the number of tasks present in an auxiliary task set.

\section{MTL Benchmarks and Leaderboards}
While there are many research efforts that evaluate MTL model performance on custom task sets, there exist several gold standard benchmarks which enable comparative evaluation. The first of these is the NLP Decathlon \citep{decanlp18}, decaNLP, which combines ten common NLP tasks/datasets: Question Answering, Machine Translation, Summarization, Natural Language Inference, Sentiment Analysis, Semantic Role Labeling, Zero-Shot Relation Extraction, Goal-Oriented Dialog, Semantic Parsing and Pronoun Resolution. Each task is assigned a scoring metric between 0 and 100. An overall decaScore is computed as the sum of all the task scores with the highest possible being 1,000. Using the Context Tasking technique, every task is represented as a natural language question, a context and an answer. The decaNLP leaderboard presents an opportunity for MTL researchers to assess model performance.

One of the most popular evaluation benchmarks used for TL and MTL alike is the General Language Understanding Evaluation (GLUE) \citep{glue18}. GLUE challenges models to solve the following 9 NLP tasks: Grammatical Acceptance Prediction, Sentiment Analysis, Paraphrasing, Semantic Equivalence, Semantic Similarity, Question Answering, Pronoun Resolution and two different Textual Entailment tasks. Each task has a defined scoring metric to evaluate task-specific performance; F1 score is commonly used among the tasks. GLUE does not require that all tasks be solved by the same model, and, as such, many top solutions have a fine-tuned model per task. An overall GLUE score, the macro-average of all tasks, is computed for each model.

In order to keep challenging researchers to push the state-of-the-art, an additional NLP benchmark, called SuperGLUE, is presented which is designed to be significantly more difficult \citep{superglue19}. The following 7 tasks are included in the SuperGLUE: Binary Question Answering, Imbalanced 3-Class Textual Entailment, Logical Causal Relationship, Textual Entailment, Binary Word-Sense Disambiguation, Pronoun Resolution and two different Multiple-Choice Question Answering tasks. Textual Entailment and Pronoun Resolution are the only two tasks from the original GLUE benchmark retained in SuperGLUE. These tasks were kept because they still showed room for improvement and proved to be two of the hardest tasks in GLUE.

\section{MTL Solutions for NLP}
There is a rich library of research presenting technical implementations and use cases for MTL models and architectures. This section provides an overview of recent state-of-the-art approaches. Table \ref{param_comparison} shows a comparison of model sizes and scores on common benchmarks.

\begin{table}[t]
	\caption{MTL Model Comparison}\smallskip
	\centering
	\resizebox{.95\columnwidth}{!}{
		\smallskip\begin{tabular}{l|l|l|l|l}
			Model & Params & GLUE & SuperGLUE & decaScore\\ \hline
			MQAN & 29M & - & - & 609.0\\
			MT-DNN & 350M & 87.6 & - & -\\
			BERT$_{Base}$ & 110M & 78.3 & - & -\\
			BERT$_{Large}$ & 340M & 80.5 & 69.0 & -\\
			BERT with PALs & 125M & - & - & -\\
			BERT+BAM & 335M & 82.3 & - & -\\
			RoBERTa & 375M & 88.5 & 84.6 & -\\
			ALBERT$_{xxl}$ Ensemble & 235M & 89.4 & - & -\\
			GPT-2 & 1,542M & - & - & -\\
			XLNet-Large & 340M & 88.4 & - & -\\
			T5-11B & 11B & 89.7 & 89.3 & -\\
		\end{tabular}
	}
	\label{param_comparison}
\end{table}

\subsection{Multi-task Question Answering Network}
The Multi-task Question Answering Network (MQAN), \citep{decanlp18}, is a natural language Context Tasking network designed to jointly learn over all tasks with no task specific weights or parameters in the network. All inputs and tasks are modeled as natural language questions and outputs in the form of a natural language answer. This enables the network to learn to solve tasks which traditionally have different input and output structures, such as machine translation and relation extraction. The authors show that MQAN is able to achieve performance comparable to ten single-task networks with no fine-tuning or task specific layers. Due to the common contextualized input design, MQAN is able to do zero-shot training and can even adapt to unseen classes in classification.

\subsection{BERT and Related Models}
Arguably one of the most important models recently proposed is BERT: Bidirectional Encoder Representations from Transformers \citep{bert19}. BERT pre-trains a transformer model \citep{transformer17} with an unsupervised multi-task objective. This pre-training objective trains the network to predict a random mask of hidden words in a text document and to predict if a shown sentence is the logical next sentence in the document via a binary classifier. Along with the novel pre-training objective, BERT also presents a mechanism for contextualizing on both the left and right text directions while other popular models, such as GPT \citep{gpt18}, are unidirectional. BERT scored competitively on the GLUE leaderboard and provided a base for researchers to build upon.

Since the release of BERT, there have been a number of modifications which have surpassed the baseline score on GLUE \citep{bertonstilts18, spanbert19,albert19}. BERT and PALs train a single BERT model to be used for all tasks jointly, as opposed to building a fine-tuned model for each task \citep{bertandpals19}. \cite{bam19} approach MTL with BERT from a different angle by doing multi-task fine-tuning through knowledge distillation and a multi-teacher paradigm, called BAM. The model is trained with a teacher annealing curriculum that gradually transfers the target learner from distillation through the teachers to a supervised MTL signal. RoBERTa is an optimized take on BERT that finds techniques which significantly improve performance \citep{roberta19}. ALBERT replaces the next sentence prediction task, proven ineffective by \cite{xlnet19}, with a sentence-order prediction pre-training task \citep{albert19}.

One notable extension of BERT with true multi-task learning across all 9 GLUE tasks is the Multi-Task Deep Neural Network (MT-DNN) \citep{mtdnn19}. The authors argue that MT-DNN has better domain transfer across tasks than standard BERT. The process begins with the regular BERT pre-training, followed by multi-task training with hard parameter sharing and a random round robin curriculum and finally ends with task-specific fine-tuning.

\subsection{GPT/GPT-2}
BERT is not the only type of language model that has successfully performed in MTL environments. GPT \citep{gpt18} is based on a multi-layer transformer network and GPT-2 \citep{gpt19} extends this model with an unsupervised multi-task pre-training objective. In inference settings GPT-2 is first task-conditioned to solve the desired task. This zero-shot type of learning can outperform the current state-of-the-art on a majority of NLP tasks. GPT-2 is also shown to perform competitively when used in a traditional pre-training and fine-tuning process. The authors have indicated that in future work they plan to assess GPT-2 performance on decaNLP and GLUE benchmarks.

\subsection{XLNet}
XLNet is proposed as a next-generation model which is intended to leverage the best features found in BERT and GPT while overcoming their intrinsic shortcomings \citep{xlnet19}. The authors claim that BERT suffers from a pre-train/fine-tune discrepancy due to the masked words introduced in pre-training. While the masked words are helpful for building a latent understanding of language, masked words are never seen in practice and thus there is a distinct difference in the training data and real-world inputs. While this simplification has worked well for BERT, \cite{xlnet19} attempt to improve performance by estimating the joint probability of the words seen in a piece of text. The authors also empirically show that BERT's next sentence prediction pre-training objective did not improve model performance and, hence, was dropped from the XLNet pre-training regimen.

\subsection{T5}
T5 \citep{t519} (Text-to-Text Transfer Transformer) is a refinement to the traditional transformer which boasts an unsupervised pre-training corpus of roughly 750 GB and uses natural language Context Tasking. The highest performing model designed by the authors contains 11 billion parameters, far more than what any other model has considered, and has beaten all other models addressed above on the GLUE and SuperGLUE leaderboards. This work provides convincing evidence regarding the claim that model capacity is an important factor in transfer learning and MTL in NLP.

\section{Current Challenges and Opportunities}
Most challenges that are still faced today in MTL are the same challenges that have existed for the past two decades. \cite{mtl97} proved that some inductive bias can hurt, and while it is still generally believed that task relatedness leads to good bias, there is no strong general notion of measuring this \citep{notion08,mtloverview17}. \cite{whichtask19} begin to address this by confronting the underlying challenge of crosstalk, in which MTL suffers from complex and competing objectives. Additional studies have researched task relationship and performance on earlier model generations, such as bi-LSTMs \citep{bentask17,mts2sl15,whenmtl17}. Studies applying similar in-depth analysis to the most recent multi-task benchmarks with the latest transformer-based models are prime research opportunities to understand better the tasks to solve and the implications of the selected models.

\cite{whichtask19} present several interesting claims which are worth exploring and applying to known MTL benchmarks. The first is that it could be better to train dissimilar tasks as opposed to semantically similar tasks. Additionally they argue that MTL performance estimates can be made by averaging the results of lesser-order task pairs. Both claims present research opportunities that could lead to better understanding of the impact of auxiliary task selection. The new set of MTL deep learning models should also be explored through probing in a manner similar to that of \cite{probing19} to understand better the impact of NLP task selection. There still is a need for deeper and more general techniques for task selection and task assessment. As research dives deeper into the implications of MTL it is important to continue strengthening the current understanding of task relationship and selection.

Curriculum learning is continuing to gain popularity and will likely become of larger interest with the introduction of standardized MTL benchmarks. Curriculum learning has not been explored much in NLP or MTL, however, it has a rich history in reinforcement learning (RL) where curriculum is used to guide trained agents to more complex and realistic behaviors \citep{curriculum17,soid17}. The curriculum is often generated in RL settings and it would be interesting to expand on these capabilities for MTL curriculum generation. These generations could leverage some form of relatedness \citep{mtdnn19} or be driven by unsupervised or latent signals \citep{task2vec19}. Other research into lifelong learning and continuous learning \citep{neverending18,continuallearning19} present new ideas and paradigms which are related to MTL and can be utilized to help solve the MTL tasks mentioned in this survey.

Although many unsupervised natural language understanding tasks have recently been used in a pre-training setting, \cite{mts2sl15} pose the question of how unsupervised objectives may impact MTL performance as auxiliary tasks. Building off the TL process there are open questions on how an MTL model can leverage the same unsupervised datasets. They argue that an auxiliary task must be compatible with the target task, and both intrinsic (perplexity) and extrinsic (accuracy) metrics must be improved on the target task when trained with the auxiliary task. \cite{whenmtl17} pose an additional question: most auxiliary tasks are classification tasks, how do regression tasks fare as auxiliary tasks? \cite{readability18} provide an example of this with text readability prediction and an auxiliary gaze prediction task. They showed that they only needed small samples from the auxiliary task, the selection of the auxiliary task was robust to small changes in the domain and the shared feature representation provably enhanced model performance. This work shows that further research into regression auxiliary tasks could help to advance MTL state-of-the-art. Finally, \cite{mtdnn19} present a unique opportunity to study how MTL architectures perform against adversarial tasks which could potentially lead to a new set of hardened auxiliary tasks. We hypothesize that Domain Regularization or Multi-Task Feature Learning could help machine learning models better withstand adversarial attacks.

Most recent advancements in TL and MTL are based off hard parameter sharing. How do model architectures, such as the transformer, perform when regularized with an MTL-based soft parameter sharing? How would this compare to standard models such as BERT and GPT and what other techniques can be borrowed from \cite{mtloverview17} for the latest generation of deep learning models?

Lastly, the biggest challenge faced in current MTL research is that fine-tuned single-task models consistently outperform non-fine-tuned MTL models that share layers \citep{decanlp18,bam19}. MTL pre-training followed by single-task fine-tuning is able to leverage the rich knowledge acquired through inductive bias, but the impact of the strong supervised signal creates narrow experts which are able to outperform the generalized experts produced by MTL. While this is fine for narrow systems designed to solve problems with expansive training datasets, this gap needs to be closed to improve performance on data sparse tasks and domains. A long-term goal that will continue to persist is to develop general experts which can compete with their single-task counterparts \citep{generalai19}.

Ultimately we find this ambitious task before us. To find ways to build robust and capable MTL models and help to enable the next generation of general Artificial Intelligence.

\bibliography{mtl_2020}

\end{document}